\documentclass[letterpaper]{article} 
\usepackage{aaai2026}
\usepackage{times}  
\usepackage{helvet}  
\usepackage{courier}  
\usepackage[hyphens]{url}  
\usepackage{graphicx} 
\usepackage{natbib}  
\usepackage{caption} 
\usepackage{algorithm}
\usepackage{algorithmic}

\usepackage{newfloat}
\usepackage{listings}
\DeclareCaptionStyle{ruled}{labelfont=normalfont,labelsep=colon,strut=off} 
\floatstyle{ruled}
\newfloat{listing}{tb}{lst}{}
\floatname{listing}{Listing}
\usepackage{xspace}
\usepackage{amsmath}
\usepackage{amssymb}
\usepackage{booktabs}
\usepackage{multirow}
\usepackage[breakable,skins]{tcolorbox}
\usepackage{lipsum}

\newcommand{\mname}{{\sc EAGLE}\xspace}  

\usepackage{helvet}
\usepackage{courier}
\usepackage{xcolor}
\newcounter{checksubsection}
\newcounter{checkitem}[checksubsection]

\title{Enhancing Uncertainty Estimation in LLMs \\
with Expectation of Aggregated Internal Belief}

\author{
    Zeguan Xiao$^{1}$\thanks{Equal contribution.},
    Diyang Dou$^{1}$\footnotemark[1],
    Boya Xiong$^{1}$\footnotemark[1],
    Yun Chen$^{1,2}$\thanks{Corresponding Authors.},
    Guanhua Chen$^{3}$\footnotemark[2]\\
}
\affiliations{
    $^1$Shanghai University of Finance and Economics;\\
    $^2$MoE Key Laboratory of Interdisciplinary Research of Computation and Economics,\\
    Shanghai University of Finance and Economics; \\
    $^3$Southern University of Science and Technology
}

\usepackage{bibentry}

\begin{document}

\maketitle

\begin{abstract}
Large Language Models (LLMs) have achieved remarkable success across a wide range of natural language tasks, but often exhibit overconfidence and generate plausible yet incorrect answers. This overconfidence, especially in models undergone Reinforcement Learning from Human Feedback (RLHF), poses significant challenges for reliable uncertainty estimation and safe deployment.
In this paper, we propose \mname (Expectation of AGgregated internaL bEief), a novel self-evaluation-based calibration method that leverages the internal hidden states of LLMs to derive more accurate confidence scores.
Instead of relying on the model's final output, our approach extracts internal beliefs from multiple intermediate layers during self-evaluation. By aggregating these layer-wise beliefs and calculating the expectation over the resulting confidence score distribution, \mname produces a refined confidence score that more faithfully reflects the model's internal certainty. Extensive experiments on diverse datasets and LLMs demonstrate that \mname significantly improves calibration performance over existing baselines.
We also provide an in-depth analysis of \mname, including a layer-wise examination of uncertainty patterns, a study of the impact of self-evaluation prompts, and an analysis of the effect of self-evaluation score range.
Our code is public at \url{https://github.com/sustech-nlp/EAGLE}.
\end{abstract}

\section{Introduction}

Large Language Models (LLMs) have demonstrated impressive capabilities across a wide spectrum of natural language understanding and generation tasks \cite{Brown2020LanguageMA, wolf2020huggingfacestransformersstateoftheartnatural, Achiam2023GPT4TR, Touvron2023LLaMAOA}. Their ability to generate coherent and contextually relevant text has led to their increasing deployment in various real-world applications. However, a critical challenge remains: LLMs often produce plausible-sounding but incorrect answers, a phenomenon known as hallucination \cite{Bai2022ConstitutionalAH}.
This challenge is further exacerbated by the lack of reliable uncertainty estimation for the outputs of LLMs that have undergone Reinforcement Learning from Human Feedback (RLHF) \cite{kadavath2022language,Tian2023JustAF}, undermining their trustworthiness and limits their safe deployment. To mitigate this issue, recent research has introduced calibration techniques, which aim to ensure that LLMs' confidence of predictions accurately reflect their likelihood of correctness \cite{kadavath2022language,Chen2023QuantifyingUI,xiong2023can}.

Among the various research directions, a particularly important approach is leveraging the LLMs' inherent capabilities for self-reflection and self-evaluation as a means of uncertainty estimation \cite{Lin2022TeachingMT, kadavath2022language, Tian2023JustAF, Chen2023QuantifyingUI, Huang2024CalibratingLG}. These methods often involve prompting the LLM to explicitly assess the uncertainty and/or correctness of its own generated responses, usually by eliciting verbalized confidence scores.

The motivation of our method stems from the observation that LLMs' internal representations, particularly the hidden states at different layers, contain rich information about LLMs' internal belief and can be used to reveal truthfulness of statements, knowledge discovery and behavior monitoring \cite{burns2022discovering,azaria2023internal,zou2023representation, laibeyond}. In particular, concurrent work \cite{ji2025calibrating} demonstrates that intermediate-layer hidden states naturally separate high-confidence from low-confidence model predictions, suggesting that the model’s latent space already encodes information about its own uncertainty. Inspired by this, we propose a novel self-evaluation-based calibration method, \mname (Hidden Layer Aggregation), for uncertainty estimation. Unlike previous methods that rely on superficial verbalized confidence scores, our approach derives the confidence score by leveraging the internal representations of LLMs.

Specifically, \mname first begins by prompting the LLM to assess the uncertainty of its own answer by providing a numerical confidence score. It then extract the hidden representations corresponding to the self-evaluation token from multiple layers of the model. These hidden states are projected into the vocabulary space to obtain layer-wise logits, which are subsequently aggregated by averaging to form a refined logit vector. By applying a softmax over the logits associated with the candidate confidence scores, \mname produces a probability distribution that reflects the model's internal belief about its answer's correctness. The final confidence score is computed as the expectation over this distribution. This layer-wise aggregation strategy enables \mname to capture richer and more nuanced confidence signals than methods based solely on the model's final output, leading to improved calibration performance across diverse tasks and model architectures.

The main contributions of this work are as follows:
\begin{itemize}
    \item We propose a novel training-free self-evaluation-based uncertainty estimation method, \mname, which leverages layer-wise hidden states to enhance calibration in LLMs.
    \item We conduct extensive empirical evaluations on three diverse datasets with state-of-the-art LLMs, showing that our proposed methods achieve superior calibration performance compared to existing baselines.
    \item We provide an in-depth analysis of \mname, including a layer-wise examination of uncertainty patterns, a study of the impact of self-evaluation prompts, and an analysis of the effect of self-evaluation score range.
\end{itemize}

\begin{figure*}[htbp]
    \centering
    \includegraphics[width=0.95\textwidth]{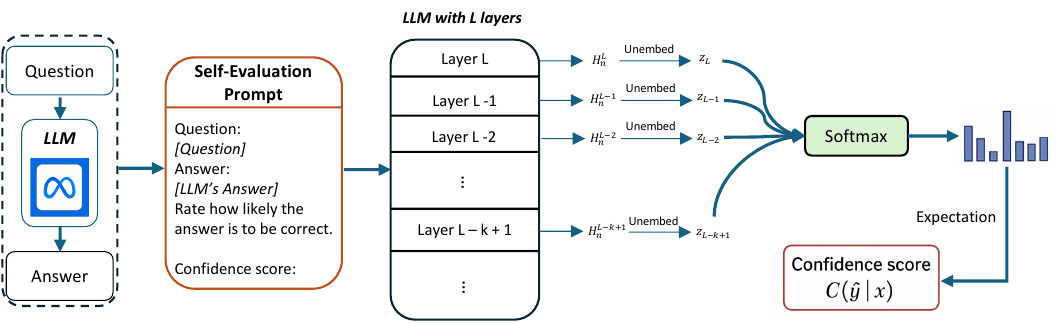}
    \caption{The \mname pipeline. First, the LLM generates an answer. Then, \mname prompts the LLM to self-evaluate the correctness of its answer, extracts the hidden states corresponding to the self-evaluation token from multiple layers, projects these hidden states to logits, and aggregates them. Finally, a softmax function is applied to the aggregated logits to compute the final confidence score.}
    \label{fig:flow}
\end{figure*}

\section{Related Work}

\paragraph{Language Models Calibration.}
Model calibration—the alignment of a model's predicted probabilities with the true likelihood of correctness \cite{Hendrycks2016ABF, Guo2017OnCO}—is crucial for user trust and informed decision-making in AI systems. In traditional supervised learning, post-hoc calibration techniques such as temperature scaling \cite{Guo2017OnCO}, Platt scaling \cite{platt1999probabilistic}, and isotonic regression \cite{niculescu2005predicting} are used to improve the reliability of model confidence estimates. However, directly applying these methods to LLMs, particularly those undergone RLHF \cite{kadavath2022language, Tian2023JustAF}, poses significant challenges.

Calibration in LLMs has been approached primarily through two lines of research.
The first involves analyzing the diversity of multiple generated outputs, where techniques such as clustering-based metrics are used to examine the variance in the model's responses \cite{kuhn2023semantic, lin2024contextualized}.
While effective in some scenarios, these methods are often computationally intensive due to the need for multiple generations per input, and the diversity observed does not always reliably indicate the correctness of individual outputs.
A second, increasingly prominent direction leverages the self-evaluation capabilities of LLMs. In this approach, the model is prompted to assess the quality or correctness of its own generated responses, typically by providing a verbalized confidence score or categorical judgment \cite{Lin2022TeachingMT, kadavath2022language, Tian2023JustAF, Chen2023QuantifyingUI, Huang2024CalibratingLG}. Self-evaluation methods offer a more direct means of accessing the model's internal assessment of its outputs, and have shown promise for improving calibration. However, most existing self-evaluation approaches rely on superficial verbalized confidence scores, which may fail to capture the nuanced confidence signals present in the model's internal representations.

\paragraph{Internal Representations of LLMs.}
Analyzing the internal representations of LLMs has emerged as a crucial direction for understanding their inner workings and decision-making processes.
A promising line of research involves interpreting these hidden states by projecting them back into the vocabulary space. This technique allows researchers to treat the representation at each layer as a distribution over the vocabulary, offering insights into the model's predictions and internal "beliefs" at different stages of processing \cite{nostalgebraist2020interpreting, Geva2022TransformerFL}.
Recent studies have leveraged this approach to probe the model's internal states for various purposes, such as discovering latent knowledge \cite{burns2022discovering}, assessing the truthfulness of statements \cite{azaria2023internal}, and even steering model behavior through representation engineering \cite{zou2023representation}. These works collectively demonstrate that a model's internal representations often contain more nuanced and faithful information than its final output layer alone.

Our work, \mname, is directly inspired by and builds upon this line of inquiry. While previous research has successfully used internal states to uncover a model's knowledge or factual alignment, we are the first to apply this to the problem of uncertainty calibration in a self-evaluation context. Unlike methods that rely on the final verbalized confidence, \mname posits that a more robust and genuine measure of the model's certainty can be derived by aggregating the logits from multiple layers. This approach allows us to tap into the rich, multi-layered confidence signals within the model, moving beyond superficial verbalized confidence score to achieve a more reliable estimation of uncertainty.

\section{Preliminaries}

Current mainstream LLMs are decoder-only transformers, which consist of a sequence of identical transformer decoder blocks stacked in series. Formally, a decoder-only transformer can be represented as a composition of functions:

$$f_{\text{transformer}} = f_{\text{unembed}} \circ \underbrace{f^{(L)} \circ f^{(L-1)} \circ \cdots \circ f^{(1)}}_{\text{L transformer layers}} \circ f_{\text{embed}}$$
where $f_{\text{embed}}: \mathbb{R}^{n} \rightarrow \mathbb{R}^{n \times d}$ is the input embedding function that maps a sequence of $n$ tokens to their corresponding $d$-dimensional embeddings, $f^{(l)}$ represents the $l$-th transformer decoder layer, and $f_{\text{unembed}}: \mathbb{R}^{n \times d} \rightarrow \mathbb{R}^{n \times |V|}$ is the output projection that transforms the final hidden states to logits over the vocabulary $V$.
The hidden state at layer $l$ is denoted as $H^{(l)} \in \mathbb{R}^{n \times d}$, where each row $H^{(l)}_i \in \mathbb{R}^{d}$ represents the hidden representation of the $i$-th token at layer $l$. This hierarchical processing allows the model to progressively refine its understanding of the input sequence, with different layers capturing varying levels of linguistic and semantic information.

\section{Methodology}

\subsection{Problem Definition}
In this work, we focus on the problem of uncertainty estimation for LLMs, which is particularly challenging for LLMs that have undergone RLHF~\cite{kadavath2022language,Tian2023JustAF}.
A key concept in uncertainty estimation is calibration. A model is considered well-calibrated if its confidence of predictions accurately reflect their likelihood of correctness.
Formally, for a given input $x$, its golden label $y$, and model prediction $\hat{y}$, a perfectly calibrated model satisfies:
\begin{equation}
\operatorname{Pr}[\hat{y} = y | C(\hat{y}|x) = p] = p, \quad \forall p \in [0,1],
\end{equation}
where $C(\hat{y}|x)$ is the confidence score assigned by the model to its prediction $\hat{y}$ given input $x$.

The most commonly used metric for quantifying the degree of miscalibration is Expected Calibration Error (ECE). ECE is defined as $\mathbb{E}[|\operatorname{Pr}(\hat{y} = y \mid C(\hat{y}|x) = p)-p|]$, which measures the difference between confidence and accuracy.
An empirical estimate of ECE is calculated by partitioning $N$ samples into $M$ bins $\{B_1, B_2, \ldots, B_M\}$ according to the confidence score $C$ predicted by the model. The ECE is then formulated as:
\begin{equation}
\label{eq:ece}
    \text{ECE} = \frac{1}{N}\sum_{m=1}^M |B_m|\left|\text{acc}(B_m) - \text{conf}(B_m)\right|,
\end{equation}
where $\text{acc}(B_m)$ and $\text{conf}(B_m)$ denote the average accuracy and confidence within bin $B_m$, respectively.
A smaller ECE indicates better calibration performance of the model.

\subsection{Motivation}

Our motivation stems from a growing body of work demonstrating that the internal representations of LLMs, specifically the hidden states across different layers, encode a wealth of information that is not always reflected in the final output. Research has shown that these internal states can be probed to discover latent knowledge, track the truthfulness of statements, and monitor the model's behavior \cite{burns2022discovering, azaria2023internal, zou2023representation, laibeyond}.
For instance, \citet{burns2022discovering} found that a simple probe applied to hidden states could consistently reveal a model's latent knowledge, even when the model's final output was incorrect. Similarly, \citet{zou2023representation} demonstrated that representations of concepts could be identified and steered within the model.
In particular, concurrent work \cite{ji2025calibrating} demonstrates that intermediate layer hidden states can separate high-confidence from low-confidence model predictions, suggesting that the model’s latent space already encodes information about its own uncertainty.

Inspired by these empirical findings, we hypothesize that by aggregating information directly from hidden states of intermediate layers, we can construct a more robust and better-calibrated uncertainty estimate. Furthermore, we posit that relying on a single point estimate for the confidence score, such as selecting the most probable score, may fail to capture the full extent of the model's uncertainty, as the entire probability distribution over possible scores contains richer information about the model's internal belief state. Based on these insights, our approach, which we call \mname, taps into the deeper, multi-layered internal reasoning of the model, bypassing the final layer's potential for overconfident posturing. It then uses the expectation of the confidence score distribution to capture a more nuanced and faithful measure of the model's internal belief.

\subsection{\mname: Expectation of Aggregated Internal Belief}

Given a question $x$ and the LLM's generated answer $\hat{y}$, we prompt the LLM to evaluate its own answer. The self-evaluation prompt is designed to elicit a confidence score about the correctness of the answer.

After the forward pass of the LLM, we obtain the hidden states $H^{(l)} \in \mathbb{R}^{n \times d}$ at each layer, where $H^{(l)}_n \in \mathbb{R}^{d}$ denotes the hidden state corresponding to the token used to predict the confidence score in each layer.
Rather than relying solely on the hidden states from the final layer to decode a confidence score ($f_{\text{unembed}} \circ H^{(L)}_n$), we instead take into account the information from all of the last $k$ layers to obtain a more nuanced confidence estimate.
The rationale behind this approach is that different layers capture varying levels of information and reasoning, which can provide a richer understanding of the model's confidence in its self-evaluation.

\begin{tcolorbox}[enhanced, colback=white, title={Self-Evaluation Prompt}]
{\small
Question: \\
\texttt{[QUESTION]} \\
Answer: \\
\texttt{[LLM'S ANSWER]} \\
Rate how likely the answer is to be correct using a number from 0 to 9:\\
- 0 = Very likely incorrect\\
- 5 = Uncertain or partially correct\\
- 9 = Very likely correct \\

Confidence score (0-9):
}
\end{tcolorbox}

Concretely, for each of the $k$ selected layers $l$, we map the hidden state $H^{(l)}_n$ to a logit $z^{(l)}_n \in \mathbb{R}^{|V|}$:
\begin{equation}
z^{(l)}_n = f_{\text{unembed}} \circ H^{(l)}_n.
\end{equation}

Subsequently, we weight and aggregate these logits from different layers:
\begin{equation}
    z_n = \sum_{l=1}^{k} w_l \cdot z^{(l)}_n,
\end{equation}
where $w_l$ is the weight assigned to the logits from layer $l$. In our implementation, we set $w_l = 1/k$ for all layers, meaning each layer contributes equally to the final logit.

Afterwards, we apply the softmax function to the aggregated logits $z_n$, considering only the logits corresponding to the confidence score tokens. Specifically, we restrict the calculation to the set $S = \{0, 1, 2, \ldots, 9\}$, which represents all possible confidence scores. This yields a probability distribution over the candidate confidence scores:

\begin{equation}
    w_s = \frac{\exp(z_{n,s})}{\sum_{s' \in S} \exp(z_{n,s'})}.
\end{equation}

Rather than merely selecting the confidence score with the highest probability like previous works \cite{Lin2022TeachingMT, kadavath2022language, Tian2023JustAF, Chen2023QuantifyingUI, Huang2024CalibratingLG}, we utilize the entire distribution to derive a more robust estimate of the model's confidence. This approach allows us to capture the nuanced uncertainty in the model's self-evaluation, avoiding reliance on a single point estimate that may not fully represent the model's internal belief. Formally, the final confidence score, $C(\hat{y}|x)$, is calculated as the examination of the confidence score distribution:

\begin{equation}
    C(\hat{y}|x) = \sum_{s \in S} w_s \cdot s
\end{equation}

In summary, \mname introduces two key innovations for self-evaluation-based calibration: (1) it aggregates hidden states from multiple layers, rather than relying solely on the final layer, to capture richer and more robust confidence signals; and (2) it computes the final confidence score as the expectation over the entire confidence score distribution, rather than selecting only the most probable score. Together, these strategies enable \mname to more faithfully reflect the model's internal belief about its answer's correctness.

\section{Experiments}

\begin{table*}[htbp]
  \centering
  \small
  \begin{tabular}{@{}llc|ccccc|ccccc@{}}
    \toprule
    \multirow{2}{*}{\textbf{Metric}} & \multirow{2}{*}{\textbf{Dataset}} & & \multicolumn{5}{c|}{\textbf{Llama3 8B}} & \multicolumn{5}{c@{}}{\textbf{Llama3 70B}} \\
    \cmidrule(lr){4-8} \cmidrule(lr){9-13}
    & & & \textbf{SE} & \textbf{Self-cons.} & \textbf{P(true)} & \textbf{CSL} & \textbf{\mname} & \textbf{SE} & \textbf{Self-cons.} & \textbf{P(true)} & \textbf{CSL} & \textbf{\mname} \\
    \midrule
    \multirow{4}{*}{ECE (↓)} 
    & TriviaQA & & 15.5 & 27.7 & 23.3 & 28.7 & \textbf{1.7} & 19.3 & 17.1 & 16.0 & 18.1 & \textbf{2.0} \\
    & GSM8K & & 17.1 & 25.4 & 25.1 & \textbf{6.3} & 7.6 & 16.2 & 5.0 & 7.5 & 7.5 & \textbf{4.9} \\
    & MMLU & & 5.1 & 7.3 & 37.9 & 39.2 & \textbf{0.4} & 8.8 & 5.0 & 21.5 & 22.2 & \textbf{2.4} \\
    & Average &  & 12.6 & 20.1 & 28.8 & 24.7 & \textbf{3.2} & 14.8 & 9.0 & 15.0 & 15.9 & \textbf{3.1} \\
    \midrule
    \multirow{4}{*}{AUROC (↑)} 
    & TriviaQA & & 60.0 & 55.7 & 60.6 & 53.0 & \textbf{61.5} & 66.2 & 59.1 & 63.0 & 52.0 & \textbf{70.4} \\
    & GSM8K & & 62.1 & 51.5 & 67.5 & 67.7 & \textbf{68.3} & 58.9 & 63.3 & 55.3 & 50.0 & \textbf{64.7 }\\
    & MMLU & & 54.9 & 54.8 & 53.3 & 50.0 & \textbf{55.1} & 55.5 & 55.5 & 53.2 & 50.0 & \textbf{57.1} \\
    & Average &  & 59.0 & 54.0 & 60.5 & 56.9 & \textbf{61.6} & 60.2 & 59.3 & 57.2 & 50.7 & \textbf{64.1} \\
    \midrule
    \multirow{2}{*}{\textbf{Metric}} & \multirow{2}{*}{\textbf{Dataset}} & & \multicolumn{5}{c|}{\textbf{Qwen2.5 7B}} & \multicolumn{5}{c@{}}{\textbf{Qwen2.5 72B}} \\
    \cmidrule(lr){4-8} \cmidrule(lr){9-13}
    & & & \textbf{SE} & \textbf{Self-cons.} & \textbf{P(true)} & \textbf{CSL} & \textbf{\mname} & \textbf{SE} & \textbf{Self-cons.} & \textbf{P(true)} & \textbf{CSL} & \textbf{\mname} \\
    \midrule
    \multirow{4}{*}{ECE (↓)} 
    & TriviaQA & & 31.6 & 32.1 & 31.2 & 67.4 & \textbf{22.1} & 18.6 & 13.8 & 13.9 & 14.5 & \textbf{1.7} \\
    & GSM8K & & 9.8 & 16.8 & 9.5 & 15.5 & \textbf{3.0} & 16.0 & 9.7 & 7.5 & 4.5 & \textbf{1.2} \\
    & MMLU & & 18.1 & 22.5 & 24.4 & 36.6 & \textbf{16.8} & 23.9 & 22.0 & 22.3 & 18.1 & \textbf{15.7} \\
    & Average &  & 19.8 & 23.8 & 21.7 & 39.8 & \textbf{14.0} & 19.5 & 15.2 & 14.6 & 12.4 & \textbf{6.2} \\
    \midrule
    \multirow{4}{*}{AUROC (↑)} 
    & TriviaQA & & 52.3 & 51.0 & 62.2 & 52.1 & \textbf{64.9} & 53.5 & 54.9 & 54.0 & 52.0 & \textbf{70.4} \\
    & GSM8K & & 54.3 & 49.7 & \textbf{74.0} & 71.7 & 72.8 & 60.1 & 64.1 & 57.0 & 53.0 & \textbf{66.5} \\
    & MMLU & & 69.7 & 69.9 & 70.3 & 51.5 & \textbf{73.9} & 53.8 & 54.3 & 54.0 & 51.9 & \textbf{57.4} \\
    & Average &  & 58.8 & 56.9 & 68.8 & 58.4 & \textbf{70.5} & 55.8 & 57.8 & 55.0 & 52.3 & \textbf{64.8} \\
    \bottomrule
    \end{tabular}
    \caption{Calibration results of different uncertainty estimation methods on Llama3 and Qwen2.5 models across multiple datasets. Lower ECE and higher AUROC indicate better calibration and discrimination, respectively. All metrics are given by $\times 100$. The best results are highlighted in bold.}
    \label{tab:main_results}
\end{table*}

\subsection{Setup}

\paragraph{Datasets.}
We evaluate the calibration performance of various methods on both open-ended and multiple-choice datasets. For open-ended tasks, we use rc.nocontext subset of TriviaQA \cite{joshi2017triviaqa} and GSM8k \cite{cobbe2021trainingverifierssolvemath}. For multiple-choice tasks, we include MMLU \cite{hendryckstest2021}.
The final evaluation is conducted on the TriviaQA validation set, GSM8k test set, and MMLU test set.
For hyperparameter tuning, we use an equal-size subset of the TriviaQA and GSM8k training sets, and the official validation set for MMLU.
These datasets cover diverse domains and reasoning complexities, enabling a comprehensive assessment of calibration across different tasks.

\paragraph{Models.}
We evaluate our method on two prominent families of LLMs: Qwen2.5 \cite{qwen2.5} and Llama3 \cite{grattafiori2024llama}. Specifically, we include Qwen2.5 models of varying scales (7B and 72B) and Llama3 models (8B and 70B). These models are selected to cover a wide range of parameter sizes and architectures, enabling a comprehensive assessment of calibration performance across different model capacities. For clarity, we refer to Qwen2.5-7B-Instruct and Meta-Llama-3-8B-Instruct as Qwen2.5 7B and Llama3 8B, respectively, and adopt similar abbreviations for other model variants throughout the paper.

\paragraph{Baselines.}
We compare our method against several established approaches for confidence estimation:

\begin{itemize}
    \item \textbf{Semantic Entropy (SE)} \cite{kuhn2023semantic}: An entropy measure that clusters semantically equivalent answers before computing uncertainty, addressing linguistic variations of correct responses.
    \item \textbf{Self-consistency (Self-cons.)} \cite{chen2024quantifying}: Confidence derived from agreement among multiple sampled answers, with higher agreement indicating greater certainty.
    \item \textbf{P(True)} \cite{kadavath2022language}: The model's self-evaluation of whether its answer is correct, obtained by prompting it to classify answers as true/false.
    \item \textbf{Contextualized Sequence Likelihood (CSL)} \cite{lin2024contextualized}: A weighted sequence likelihood that uses attention mechanisms to focus on relevant tokens when computing answer probabilities.
\end{itemize}

\paragraph{Evaluation Metrics.}
To comprehensively assess the quality of confidence scores, we consider two orthogonal evaluation tasks: calibration and failure prediction.
Calibration measures how closely the predicted confidence aligns with empirical accuracy. We quantify calibration using ECE, which computes the average absolute difference between predicted confidence and observed accuracy across confidence intervals, as defined in Eq \ref{eq:ece}. For all experiments, we use 10 bins for ECE calculation.
Failure prediction, in contrast, evaluates whether the model assigns higher confidence to correct predictions and lower confidence to incorrect ones, reflecting the discriminative power of the confidence scores. For this, we report the Area Under the Receiver Operating Characteristic Curve (AUROC), which measures the ability of confidence scores to distinguish between correct and incorrect outputs. Together, ECE and AUROC provide a comprehensive evaluation of both the reliability and discriminative utility of model confidence scores.

\subsection{Main Results}

Table \ref{tab:main_results} presents the main experimental results, comparing the calibration performance of our proposed method, \mname, against several baselines\footnote{We observe that our reported CSL results differ from those presented in the original paper \cite{lin2024contextualized}. The discrepancy arises from the use of different model types: the original CSL results were obtained using base LLMs, whereas our experiments employ instruction-tuned (RLHF) models. RLHF models are known to exhibit weaker calibration compared to base LLMs \cite{Tian2023JustAF}, leading to the observed divergence.}. The results demonstrate that \mname consistently achieves superior calibration performance. 

Across all evaluated models and datasets, \mname shows a significant improvement in ECE, the primary metric for calibration. For instance, on the TriviaQA dataset, \mname reduces the ECE of the Llama3 8B model to 1.7, a substantial improvement over the best baseline (SE) which scored 15.5. Similarly, for the Llama3 70B model, \mname achieves an ECE of 2.0, far surpassing the next best baseline's 16.0. This trend of strong ECE reduction holds for the Qwen2.5 models as well, where \mname consistently delivers the lowest ECE scores, indicating a much better alignment between the model's confidence and its actual accuracy. On average, \mname achieves the lowest ECE across all settings, highlighting its effectiveness and robustness.

In addition to superior calibration, \mname also demonstrates strong discriminative power, as measured by the AUROC metric. It consistently achieves the highest or highly competitive AUROC scores, indicating its ability to effectively distinguish between correct and incorrect answers. For example, with the Llama3 70B model on TriviaQA, \mname achieves an AUROC of 70.4, which is significantly higher than all baselines. While some baselines occasionally achieve a competitive AUROC on specific tasks (e.g., P(true) on GSM8K with Qwen2.5 7B), they do so with a much poorer ECE. In contrast, \mname provides a better balance, excelling in both calibration and failure prediction.

\subsection{Ablation Study}
\label{sec:ablation_study}

\begin{table*}[htbp]
\centering
\small
\begin{tabular}{cccccc|cc|cc}
\toprule
\multicolumn{6}{c|}{\textbf{Strategies For \mname}} & \multicolumn{2}{c|}{\textbf{Llama3 8B}} & \multicolumn{2}{c@{}}{\textbf{Qwen2.5 7B}} \\
\midrule
Last layer & Last-n layers & Prob. agg. & Logits agg. & Max. & Exp. & TriviaQA & GSM8K & TriviaQA & GSM8K \\
\midrule
$\checkmark$ & $\times$ & $\times$ & $\times$ & $\checkmark$ & $\times$ & 27.9 & 25.4 & 31.6 & 9.8 \\
$\checkmark$ & $\times$ & $\times$ & $\times$ & $\times$ & $\checkmark$ & 18.8 & 22.7 & 30.5 & 8.5 \\
$\times$ & $\checkmark$ & $\checkmark$ & $\times$ & $\checkmark$ & $\times$ & 23.1 & 25.4 & 31.6 & 9.8 \\
$\times$ & $\checkmark$ & $\checkmark$ & $\times$ & $\times$ & $\checkmark$ & 63.5 & 58.2 & 64.1 & 84.6 \\
$\times$ & $\checkmark$ & $\times$ & $\checkmark$ & $\checkmark$ & $\times$ & 28.3 & 25.4 & 31.8 & 10.1 \\
$\times$ & $\checkmark$ & $\times$ & $\checkmark$ & $\times$ & $\checkmark$ & \textbf{1.7} & \textbf{7.6} & \textbf{22.1} & \textbf{3.0} \\
\bottomrule
\end{tabular}
\caption{Ablation study on the components of \mname. We compare different strategies for layer selection (last layer vs. last-n layers), aggregation method (probability vs. logits), and final score derivation (max vs. expectation). The reported metric is ECE (↓) on TriviaQA and GSM8K datasets for Llama3 8B and Qwen2.5 7B models. The best results are highlighted in bold.}
\label{tab:ablation}
\end{table*}

To dissect the components of our proposed \mname method and validate our design choices, we conduct a comprehensive ablation study. We investigate three key aspects of our framework: (1) the use of hidden layers for extracting model's internal belief (using only the final layer versus aggregating the last-$n$ layers), (2) the aggregation strategy (aggregating layer-wise probabilities versus layer-wise logits), and (3) the method for deriving the final confidence score (taking the score with the maximum probability versus computing the expectation over the score distribution). The results of these ablations are presented in Table \ref{tab:ablation}.

Our findings consistently highlight the superiority of the full \mname configuration. First, comparing the use of only the \textbf{last layer} against aggregating the \textbf{last-n layers}, we observe a marked improvement in calibration performance when information from multiple layers is integrated. This supports our core hypothesis that a more robust uncertainty signal can be obtained by tapping into the representations distributed across several layers, rather than relying solely on the final output layer.

Second, we examine the aggregation method. The results clearly indicate that \textbf{logits aggregation (Logits agg.)} significantly outperforms \textbf{probability aggregation (Prob. agg.)}. Aggregating logits before applying the softmax function preserves a richer, more nuanced signal of the model's internal confidence. In contrast, aggregating probabilities, which are post-softmax and thus normalized at each layer, leads to a loss of information and results in substantially poorer calibration.
\begin{figure}[tbp]
    \centering
    \includegraphics[width=0.45\textwidth]{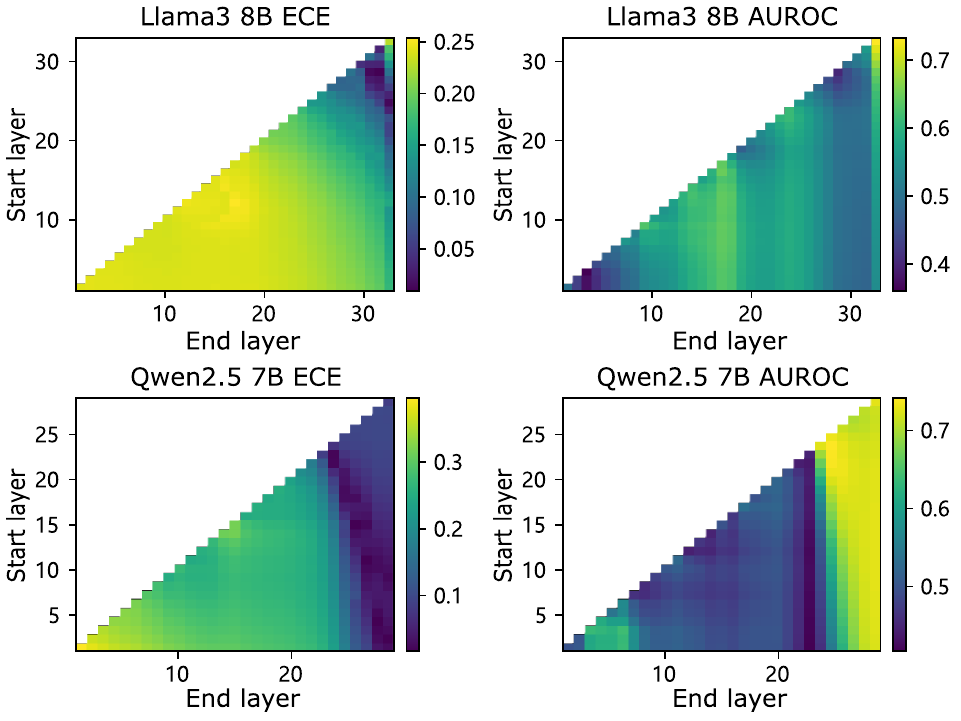}
    \caption{Layer-wise calibration performance on GSM8K for Llama3 8B and Qwen2.5 7B. The x-axis and y-axis represent the end and start layers for hidden state aggregation, respectively. Darker colors in ECE heatmaps and brighter colors in AUROC heatmaps indicate better calibration.}
    \label{fig:layer_performance}
\end{figure}

Finally, we compare two approaches for deriving the final confidence score from the aggregated distribution. Using the \textbf{expectation (Exp.)} of the score distribution consistently yields better-calibrated results than simply taking the score with the \textbf{maximum probability (Max.)}. This demonstrates that the full distribution over the confidence scores contains valuable information, and computing its expected value provides a more accurate and stable estimate of the model's true uncertainty.

\section{Analyses}

\subsection{Calibration Performance Among Layers}

To understand how different layers contribute to uncertainty estimation, we conduct a layer-wise analysis of our proposed method. We vary the range of layers from which hidden states are aggregated—from a start layer $m$ to an end layer $n$—and evaluate the resulting calibration performance. Figure \ref{fig:layer_performance} illustrates the ECE and AUROC scores for different layer combinations on the GSM8K dataset for both Llama3 8B and Qwen2.5 7B.

The heatmaps reveal a clear and consistent pattern: the calibration performance, as measured by both ECE and AUROC, is significantly better when aggregating hidden states from the later layers of the models. For both Llama3 and Qwen2.5, the optimal performance (lowest ECE and highest AUROC) is concentrated in the top-right corner of the plots. This indicates that the most reliable confidence signals are encoded in the final few layers.

While aggregating a block of layers from an intermediate start layer $m$ to an end layer $n$ can be effective, our analysis shows that simply aggregating the last few layers (i.e., a `last-n' strategy) already yields near-optimal results. The performance gain from selecting a more complex `m-n` block from earlier or middle parts of the model is marginal compared to the strong performance of the final layers. Therefore, for simplicity and efficiency, using a `last-n' aggregation strategy is a robust and sufficient choice for \mname.

\begin{table}[t]
\centering
\begin{tabular}{lll}
\toprule  
\textbf{Model} & \textbf{Method} & \textbf{Acc}\\ \midrule
\multirow{2}{*}{Llama3 8B} & Baseline & 71.3 \\ 
          & \mname            & \textbf{74.8}  \\ \midrule
\multirow{2}{*}{Qwen2.5 7B}  & Baseline & 67.5 \\
           & \mname            & \textbf{69.4} \\ \bottomrule
\end{tabular}
 \caption{\mname for answer selection. We demonstrate the accuracy of TriviaQA using different methods.}
 \label{tab:reliable_acc}
\end{table}

\subsection{\mname for Answer Selection}

One straightforward application of \mname is to apply it to select more reliable answers from LLMs. Here we conduct experiments on the TriviaQA dataset on both Qwen2.5 7B and Llama3 8B models. Specifically, we generate 5 different answers for each question with a temperature of 1.0, and use the first answer as the baseline. We then compare its accuracy with the answer selected based on the highest \mname confidence score. 
As shown in Table \ref{tab:reliable_acc}, \mname consistently improves the accuracy by selecting more reliable answers, outperforming the baseline on both models.

\subsection{Sensitivity to Prompts}

To investigate the robustness of \mname to different self-evaluation prompt formulations, we evaluate the calibration performance using two variants of self-evaluation prompts. The details of these prompt variants are shown below:

\begin{tcolorbox}[enhanced, colback=white, title={Prompt Variant A (Without score criteria)}]
{\small
Question: \\
\texttt{[QUESTION]} \\
Answer: \\
\texttt{[LLM'S ANSWER]} \\
Rate how likely the answer is to be correct using a number from 0 to 9:\\
Confidence score (0-9):
}
\end{tcolorbox}

\begin{tcolorbox}[enhanced, colback=white, title={Prompt Variant B (CoT)}]
{\small
Question: \\
\texttt{[QUESTION]} \\
Answer: \\
\texttt{[LLM'S ANSWER]} \\
Rate how likely the answer is to be correct using a number from 0 to 9:\\
- 0 = Very likely incorrect\\
- 5 = Uncertain or partially correct\\
- 9 = Very likely correct \\
You should first provide an explanation in a few sentences, then provide your score.\\
The output format should look as follows:\\
Explanation: \textless write your explanation here\textgreater \\
Confidence score (0-9): \textless give your score here\textgreater
}
\end{tcolorbox}

The results in Table \ref{tab:prompt_sensitivity} show that our method exhibits robustness to the prompt, although the choice of prompt does have an impact on performance.
Our main prompt, which provides clear criteria for the scores, generally yields the best or highly competitive results. For instance, on the GSM8K dataset, all three prompt variants produce similar ECE and AUROC scores for both models, indicating that \mname is not very sensitive in this scenario.
However, on TriviaQA, the performance varies more. The CoT prompt (Variant B) leads to a noticeable degradation in calibration for both models. This suggests that while CoT can encourage reasoning, it may also introduce noise into the self-evaluation process, which can interfere with the internal confidence signals that \mname relies on.
Overall, while the performance is not entirely independent of the prompt design, \mname demonstrates reasonable stability, particularly when provided with a clear and direct self-evaluation format.

\subsection{Effect of Confidence Score Range}

\begin{table}[t]
\centering
\small
\setlength{\tabcolsep}{5pt}
\begin{tabular}{@{}lc|cc|cc@{}}
\toprule
\multirow{2}{*}{\textbf{Model}} & \multirow{2}{*}{\textbf{Variant}} & \multicolumn{2}{c|}{\textbf{TriviaQA}} & \multicolumn{2}{c@{}}{\textbf{GSM8K}} \\
\cmidrule(lr){3-4} \cmidrule(lr){5-6}
& & \textbf{ECE} & \textbf{AUROC} & \textbf{ECE} & \textbf{AUROC} \\
\midrule
\multirow{3}{*}{Llama3 8B} 
& Main & \textbf{1.7} & \textbf{61.5} & 7.6 & \textbf{68.3} \\
& A & 4.4 & 60.1 
& 6.6 & 65.9 
\\
& B & 13.6 & 57.5 
& \textbf{5.8} & 65.0 
\\
\midrule
\multirow{3}{*}{Qwen2.5 7B} 
& Main & \textbf{22.1} & \textbf{64.9}
& 3.0& \textbf{72.8}
\\
& A & 24.5 & 60.2 
& 3.2 & 71.1 
\\
& B & 28.5 & 58.5 
& \textbf{2.5} & 72.5 
\\
\bottomrule
\end{tabular}
\caption{Comparison of prompt for \mname method. Results show ECE and AUROC performance across different prompt variants on TriviaQA and GSM8K datasets for Llama3 8B and Qwen2.5 7B models. All metrics are given by $\times 100$. Main Prompt results are taken from Table \ref{tab:main_results}.}
\label{tab:prompt_sensitivity}
\end{table}

\begin{table}[t]
\centering
\small
\begin{tabular}{@{}lcc@{}}
\toprule
\textbf{Score Range} & \textbf{ECE (↓)} & \textbf{AUROC (↑)} \\
\midrule
0 to 9 & \textbf{7.6} & \textbf{68.3}
\\
0 to 4 & 8.6 & 64.7 
\\
0 to 1 & 11.1 & 61.0 
\\
\bottomrule
\end{tabular}
\caption{Effect of confidence score range on calibration performance. Results are shown for Llama3 8B on GSM8k dataset. All metrics are given by $\times 100$.}
\label{tab:score_range}
\end{table}

To assess the impact of the confidence score's granularity on our method's performance, we conducted an experiment varying the score range in the self-evaluation prompt. We tested three different ranges: 0--9, 0--4, and 0--1. The results for the Llama3 8B model on the GSM8k dataset are presented in Table \ref{tab:score_range}.

The findings indicate a clear relationship between the score range and calibration performance. The 0--9 range yields the best results, achieving the lowest ECE (7.6) and the highest AUROC (68.3). As the range narrows, performance deteriorates. The 0--4 range shows a higher ECE and lower AUROC, and the binary 0--1 range performs the worst. This suggests that a wider, more granular scale allows the model to express its uncertainty more precisely. The fine-grained range enables \mname to capture more nuanced signals from the model's internal belief, leading to better-calibrated confidence scores.

\section{Conclusion}

In this paper, we introduced \mname, a novel elf-evaluation-based calibration method for LLMs that leverages the model's internal representations to derive a more accurate and robust uncertainty estimate. By aggregating hidden states from multiple layers and the expectation of the confidence score distribution, \mname effectively captures the model's true belief about its answer's correctness. Our extensive experiments across various models and datasets demonstrate that \mname significantly outperforms existing methods in both calibration and failure prediction tasks. The ablation study further validates the effectiveness of our design choices.

\section*{Acknowledgements}
This project was supported by National Natural Science Foundation of China (No. 62306132), Guangdong Basic and Applied Basic Research Foundation (No. 2025A1515011564) and Natural Science Foundation of Shanghai (No. 25ZR1402136). This work was done during Zeguan's internship at SUSTech. We thank the anonymous reviewers for their insightful feedback on this work.

\bibliography{aaai2026}  

\end{document}